\documentclass[11pt,a4paper]{article}
\usepackage{acl2019}
\usepackage{times}
\usepackage{latexsym}
\usepackage{soul}

\usepackage{url}
\usepackage{boxedminipage}
\usepackage{xcolor}
\usepackage{amsmath}
\usepackage{bm}
\usepackage{xspace}
\usepackage{enumitem}

\aclfinalcopy % Uncomment this line for the final submission

\usepackage{multirow}
\usepackage{booktabs}
\usepackage{array}
\newcommand{\PreserveBackslash}[1]{\let\temp=\\#1\let\\=\temp}
\newcolumntype{C}[1]{>{\PreserveBackslash\centering}p{#1}}

\usepackage{graphicx} %package to manage images
\graphicspath{ {figures/} }
\usepackage{subcaption}
\usepackage{rotating}
\usepackage{ulem}

\usepackage{tikz}
\usetikzlibrary{calc}
\usetikzlibrary{decorations.pathmorphing}
\usetikzlibrary{shapes,arrows,chains}
\usetikzlibrary{shapes.geometric}
\usetikzlibrary{patterns}
\usetikzlibrary{positioning}
\tikzset{>=latex}
\usepackage{pgfplots}

\usepackage{txfonts}

\newcommand{\jfleg}{\textsc{jfleg}\xspace}
\newcommand{\fce}{\textsc{fce}\xspace}
\newcommand{\lang}{\textsc{Lang8}\xspace}
\newcommand{\wiloc}{\textsc{WI+loc}\xspace}

\newcommand{\nucle}{\textsc{nucle}\xspace}

\newcommand{\xvec}{\mathbf{x}}
\newcommand{\yvec}{\mathbf{y}}
\newcommand{\xtvec}{\tilde{\mathbf{x}}}
\newcommand{\ytvec}{\tilde{\mathbf{y}}}

\DeclareMathOperator{\mt}{MT}

\DeclareMathOperator{\bleu}{BLEU}
\DeclareMathOperator{\nr}{NR}

\title{An Analysis of Source-Side Grammatical Errors in NMT}

\author{Antonios Anastasopoulos \\
  Language Technologies Institute\\
  Carnegie Mellon University \\
 {\tt aanastas@cs.cmu.edu}
 }
\date{}

\begin{document}
\maketitle
%\blfootnote{$^\dagger$Equal contribution.}
\begin{abstract}
The quality of Neural Machine Translation (NMT) has been shown to significantly degrade when confronted with source-side noise. 
We present the first large-scale study of state-of-the-art English-to-German NMT on real grammatical noise, by evaluating on several Grammar Correction corpora.
We present methods for evaluating NMT robustness without true references, and we use them for extensive analysis of the effects that different grammatical errors have on the NMT output.
We also introduce a technique for visualizing the divergence distribution caused by a source-side error, which allows for additional insights.
\end{abstract}

\normalem

\section{Introduction}

Neural Machine Translation (NMT) has become the \textit{de facto} option for industrial systems in high-resource settings \cite{wu2016google,humanparity,crego2016systran} while dominating public benchmarks \cite{bojar2018proceedings}. 
However, as several works have shown, it has a notable shortcoming (among others, see \citet{koehn2017six} for relevant discussion) in dealing with source-side noise, during both training and inference.
%Even under low-resource conditions, neural models were recently shown to outperform traditional statistical approaches \cite{nguyen+chiang:naacl2018}.

\citet{heigold2017robust} as well as
\citet{belinkov2017synthetic} pointed out the degraded performance of character- and subword-level NMT models when confronted with synthetic character-level noise --like swaps and scrambling-- on French, German, and Czech to English MT. \citet{belinkov2017synthetic} and \citet{cheng-EtAl:2018:Long} also studied synthetic errors from word swaps extracted from Wikipedia edits.
%\citet{sperbertoward} also shows that NMT suffers when presented with noise-ridden input, as a result of noisy speech recognition.
\citet{lui2019nmtnonnative} focused on a small subset of grammatical errors (article, preposition, noun number, and subject-verb agreement) and evaluated on English-to-Spanish synthetic and natural data.

%However, there are still several shortcomings of NMT that need to be addressed: a (non-exhaustive) list of six challenges is discussed by \citet{koehn2017six}, including out-of-domain testing, rare word handling, the wide-beam problem, and the large amount of data needed for learning. An additional challenge is robustness to noise, both during training and at inference time. 

However, no previous work has extensively studied the behavior of a state-of-the-art (SOTA) model on natural occurring data. \citet{belinkov2017synthetic} only trained their systems on about~200K parallel instances, while \citet{heigold2017robust} and \citet{lui2019nmtnonnative} trained on about~2M parallel sentences from the WMT'16 data. Importantly, though, none of them utilized vast monolingual resources through back-translation, a technique that has been consistently shown to lead to impressively better results \cite{sennrich-haddow-birch:2016:P16-11}.
%Previous work has pointed out this flaw, mainly focusing on character-level errors of the 

In this work, we perform an extensive analysis of the performance of a \textit{state-of-the-art} English-German NMT system, with regards to its robustness against real grammatical noise.
We propose a method for robustness evaluation without gold-standard translation references, and perform experiments and extensive analysis on all available English Grammar Error Correction (GEC) corpora. Finally, we introduce a visualization technique for performing further analysis.
%In this paper, we study the effect of a specific type of noise in NMT: grammatical errors.
%We primarily focus on errors that are made by non-native source-language speakers (as opposed to dialectal language, SMS or Twitter language). Not only is this linguistically important, but we believe that it would potentially have great social impact.

\section{Data and Experimental Settings}
%\subsection{Grammar Error Correction Corpora}
\label{sec:gec}

To our knowledge, there are six publicly available corpora of non-native or erroneous English that are annotated with corrections and which have been widely used for research in GEC. 

The NUS Corpus of Learner English (\textsc{NUCLE}) contains essays written by students at the National University of Singapore \cite{dahlmeier2013building}. It has become the main benchmark for GEC, as it was used in the CoNLL GEC Shared Tasks \cite{ng2013conll,ng2014conll}. 
The Cambridge Learner Corpus First Certificate in English \textsc{FCE} corpus\footnote{We use the publicly available portion.} \cite{yannakoudakis2011new} consists of essays collected from learners taking the Cambridge
Assessment's English as a Second or Other Language (ESOL) exams.\footnote{\url{https://www.cambridgeenglish.org/}}
The \textsc{Lang-8} corpus \cite{tajiri2012tense} was harvested from user-provided corrections in an online learner forum. Both have also been widely used for the GEC Shared Tasks.
%and the \textsc{AESW} 2016 Shared Task corpus, which contains corrections on texts from scientific journals.
Another small corpus developed for evaluation purposes
is the JHU FLuency-Extended GUG corpus (\textsc{JFLEG}) \cite{napoles-sakaguchi-tetreault:2017:EACLshort} with correction annotations that include extended fluency edits rather than just minimal grammatical ones.
%That way, the corrected sentence is not just grammatical, but also guaranteed to be fluent.
The Cambridge English Write \& Improve (W\&I) corpus \cite{andersen2013developing} is collected from an online platform where English learners submit text and professional annotators correct them, also assigning a CEFR level of proficiency \cite{council2001common}.
Lastly, we use a portion of the LOCNESS corpus,$^3$
%\footnote{Downloaded from \url{https://www.cl.cam.ac.uk/research/nl/bea2019st/}}
a collection of essays written by \textit{native} English speakers.~50 essays from LOCNESS were annotated by W\&I annotators for grammatical errors, so we will jointly refer to these two corpora as \wiloc.

All datasets were consistently annotated for errors with ERRANT \cite{bryant-felice-briscoe:2017:Long}, an automatic tool that categorizes correction edits.\footnote{\nucle, \lang, \fce, and \wiloc are pre-annotated with ERRANT for the \href{https://www.cl.cam.ac.uk/research/nl/bea2019st/}{BEA 2019 GEC Shared Task}. We also annotated \jfleg.} This allows us to consistently aggregate results and analysis across all datasets.

\subsection{Notation and Experimental Settings}

Throughout this work, we use the following notations:
\begin{itemize}[noitemsep]
    \item $\mathbf{x}$: the original, noisy, potentially ungrammatical English sentence. Its tokens will be denoted as $x_i$.
    \item $\tilde{\mathbf{x}}$: the English sentence with the correction annotations applied to the original sentence $x$, which is deemed fluent and grammatical. Again, its tokens will be denoted as $\tilde{x}_i$.
    \item $\mathbf{y}$: the output of the NMT system when $\mathbf{x}$ is provided as input (tokens: $y_j$).
    \item $\tilde{\mathbf{y}}$: the output of the NMT system when $\tilde{\mathbf{x}}$ is provided as input (tokens: $\tilde{y}_j$).
\end{itemize}

For the sake of readability, we use the terms grammatical errors, noise, or edits interchangeably. In the context of this work, they will all denote the annotated grammatical errors in the source sentences ($\mathbf{x}$). We also define the number of errors, or the amount of noise in the source, to be equivalent to the number of annotated necessary edits that the source $\mathbf{x}$ requires to be deemed grammatical ($\tilde{\mathbf{x}}$), as per standard GEC literature.

The main focus of our work is the performance analysis of the NMT system, so our experimental design is fairly simple. 
We use the SOTA NMT system of \citet{edunov2018} for translating both the original and the corrected English sentences for all our GEC corpora.\footnote{We use all data, concatenating train, dev, and test splits. We sample 150K sentences from \lang.} The system achieved the best performance in the WMT 2018 evaluation campaign, using an ensemble of~6 deep transformer models trained with slightly different back-translated data.\footnote{Refer to \cite{edunov2018} for further system details.}

\section{Evaluating NMT Robustness without References}

When not using human judgments on output fluency and adequacy, Machine Translation is typically evaluated against gold-standard reference translations with automated metrics like BLEU \cite{papineni2002bleu} or METEOR \cite{banerjee2005meteor}.
However, in the case of GEC corpora, we do not have access to translations -- only monolingual data (potentially with ungrammaticalities) and correction annotations.\footnote{The ideal way to potentially obtain such references of noisy text is debatable, and the extent to which humans are able to translate ungrammatical text is unknown. A well-crafted investigation could ideally elicit translations of both original and (the multiple versions of) corrected texts from multiple translators in order to study this issue. Although we highly encourage such a study, we could not conduct one due to budgetary constraints.}
Quality Estimation for MT also operates in a reference-less setting (see \citet{specia2018quality} for definitions and an overview of the field) and is hence very related to our work, but is more aimed towards towards \textit{predicting} the quality of the translation. Our goal instead, is to analyze the behavior of the MT system when confronted with ungrammatical input.
Reference-less evaluation has also been proposed for text simplification \cite{martin2018reference} and GEC \cite{napoles2016there}, while the grammaticality of MT systems' outputs has been evaluated with target-side contrastive pairs \cite{sennrich2017grammatical}.
%and with a word disambiguation suite \cite{rios2018word}.

In this work, the core of our evaluation of a system's robustness lies in the following observation:
%In order to evaluate a system's robustness, the following observation lies in the core of our analysis: 
\textbf{a perfectly robust-to-noise MT system would produce the exact same output for the clean and erroneous versions of the same input sentence.}

Denoting a perfect MT system as a function~$\mt^{perfect}(\cdot)$ over input sentences to the correct output sentences~$\hat{\mathbf{y}}$, then both input sentences~$\mathbf{x}$ and~$\tilde{\mathbf{x}}$ would yield the same output:
%{\setlength{\abovedisplayskip}{5pt}
%\setlength{\belowdisplayskip}{5pt}
\[
\hat{\mathbf{y}} = \mt^{perfect}(\mathbf{x}) = \mt^{perfect}(\tilde{\mathbf{x}}).
\]
%}
In our case,~$\hat{\mathbf{y}}$ is unknown and we only have access to a very good (but still imperfect) system~$\mt^{actual}(\cdot)$. We propose, therefore, to treat the system's output of the cleaned input ($\tilde{\mathbf{y}}$) as reference. Our assumption is that~$\tilde{\mathbf{y}}$ is a good approximation of the correct translation~$\hat{\mathbf{y}}$:
%{\setlength{\abovedisplayskip}{5pt}
%\setlength{\belowdisplayskip}{5pt}
\[
\hat{\mathbf{y}} \approx \mt^{actual}(\tilde{\mathbf{x}}) = \tilde{\mathbf{y}}.
\]
%}
Under this assumption, we can now evaluate our system's robustness by comparing $\mathbf{y}$ and $\tilde{\mathbf{y}}$ using automated metrics at the corpus or sentence level.
Here we list the metrics that we use and briefly discuss their potential shortcomings.

%\begin{itemize}[noitemsep]
    %\item 
\paragraph{Robustness Percentage (RB):} Given a GEC corpus $\{X, \tilde{X}\}$, this corpus-level metric evaluates the percentage at which the system outputs agree at the sentence level:
    \[
    RB = \frac{\sum_{\mathbf{x},\tilde{\mathbf{x}} \in \{X,\tilde{X}\}}c_{agree}(\mt(\mathbf{x}),\mt(\tilde{\mathbf{x}}))}{\vert X \vert},
    \]
    \[   
    c_{agree}(\mathbf{y},\tilde{\mathbf{y}}) = 
     \begin{cases}
       1 &\quad\text{if } \mathbf{y}=\tilde{\mathbf{y}},\\ 
       0 &\quad\text{otherwise.}\\
     \end{cases}
    \]
    %\item 
\paragraph{f-BLEU:} BLEU is the most standard MT evaluation metric, combining n-gram overlap accuracy with a brevity penalty. We calculate sentence- and corpus-level BLEU-4 scores for every $\mathbf{y}$ with $\tilde{\mathbf{y}}$ as the reference. Note that the BLEU scores that we obtain in our experiments are not comparable with any previous work (as we do not use real references) so we denote our metric as faux BLEU (f-BLEU) to avoid confusion.\footnote{In absolute numbers, we obtain higher scores than the scores of a MT system compared against actual references: the best English-German system from the WMT 2018 evaluation \cite{edunov2018} obtained a BLEU score of~46.5; our f-BLEU scores are in the [37-65] range, but we consider them informative only when viewed relative to other f-BLEU scores.}

\paragraph{f-METEOR:} Same as above, we define faux-METEOR using the METEOR MT metric \cite{denkowski2014meteor} which is more semantically nuanced than BLEU.
\begin{table*}[t]
    \centering
    \begin{tabular}{cc|cc|c|cc|c}
    \toprule
    \multicolumn{2}{c|}{\multirow{2}{*}{dataset}} & number of & average & \multirow{2}{*}{RB} & \multicolumn{2}{|c|}{over non-robust sent} & \multirow{2}{*}{NR} \\
    && sentences & \#corr/sent. & & f-BLEU & f-METEOR \\
    \midrule
    \multirow{4}{*}{\wiloc} &
    A & 9K & 3.4 & 17.77 & 46.75 &  65.29 & 2.12\\
    &B & 10K & 2.6 & 21.17 & 54.72 & 70.80 & 2.39 \\
    &C & 5.9K & 1.8 & 29.07 & 63.46 & 76.63 & 2.73 \\
    &N & 500 & 1.8 & 28.80 & 64.79 &  77.35 & 3.23 \\
    \midrule
    \multicolumn{2}{c|}{\nucle} & 21.3K & 2.0 &  20.69 & 59.97 &  74.6 & 2.92\\
    \multicolumn{2}{c|}{\fce} & 20.7K & 2.4 & 20.48 & 50.45 &  67.49 & 2.43 \\
    \multicolumn{2}{c|}{\jfleg} & 1.3K & 3.8 & 12.42 & 42.05 &  61.99 & 2.18  \\
    \multicolumn{2}{c|}{\lang} & 149.5K & 2.4 & 16.06 & 37.15 & 58.89 & 2.20 \\
    \midrule
    \multicolumn{2}{c|}{\textsc{ALL}$\setminus$\lang} & 69K & 2.4 & 20.94 & 54.65 & 70.64 & 2.55 \\
    \multicolumn{2}{c|}{\textsc{ALL}} & 218.5K & 2.4 & 17.60 & 42.65 &  62.59 & 2.55 \\
    
\bottomrule
    \end{tabular}
    \caption{Aggregate results across all datasets. As expected, the NMT system's performance deteriorates as input noise increases. For all metrics except NR, higher scores are better.}
    \label{tab:dataset}
\end{table*}

\iffalse
For all these metrics, the higher scores the better. A score of 100 would mean that the system is perfectly robust to noise (following our definition of robustness).
\fi

\paragraph{Target-Source Noise Ratio (NR):}
A notable drawback of all the previously discussed metrics is that they do not take into account the source sentences $\mathbf{x}$ and $\tilde{\mathbf{x}}$ or their distance. However, it is expected that minimal perturbations of the input (e.g. some missing punctuation) will also be minimally reflected in the difference of the outputs, while more distant inputs (which means higher levels of noise in the uncorrected source) would lead to more divergent outputs.
To account for this observation, we propose Target-Source Noise Ratio (NR) which factors the distance of the two source sentences into a metric. The distance of two sentences can be measured by any metric like BLEU, METEOR, etc. We simply use BLEU:
    \[
    \nr(\mathbf{x},\tilde{\mathbf{x}},\mathbf{y},\tilde{\mathbf{y}}) = \frac{d(\mathbf{y},\tilde{\mathbf{y}})}{d(\mathbf{x},\tilde{\mathbf{x}})} = \frac{100-\bleu(\mathbf{y},\tilde{\mathbf{y}})}{100-\bleu(\mathbf{x},\tilde{\mathbf{x}})}.
    \]
If the average (corpus-level) Noise Ratio score is smaller than 1 ($\nr(X,\tilde{X},Y,\tilde{Y}) < 1$) then we can infer that the MT system reduces the relative amount of noise, as there is higher relative n-gram overlap between the outputs than the inputs. On the other hand, if it is larger than 1, then the MT system must have introduced even more noise.\footnote{As presented, the NR metric assumes that the length of the input and target sentences are comparable. In the English-German case, this is more or less correct. A more general implementation could include a discount term based on the average sentence length ratio of the two languages.}

Recently, \citet{michel2019evaluation} proposed a criterion for evaluating adversarial attacks, which requires also having access to the correct translation $\hat{\mathbf{y}}$. Using a similarity function $s(\cdot)$, they declare an adversarial attack to be successful when:
\[
s(\xvec,\xtvec) + \frac{s(\yvec,\hat{\yvec}) - s(\ytvec,\hat{\yvec})}{s(\yvec,\hat{\yvec})} > 1
\]
In our reference-less setting, assuming $\hat{\yvec} \approx \ytvec$ leads to $s(\ytvec,\hat{\yvec})=1$. Finally, representing the similarity function with a distance function $s(\cdot) = 1 - d(\cdot)$ and simple equation manipulation, the criterion becomes exactly our Target-Source Noise Ratio.
We have, hence, arrived at a reference-less criterion for evaluating any kind of adversarial attacks.\footnote{Indeed, grammatical noise is nothing more than natural occurring adversarial noise.}

\section{Analysis}
\label{sec:analysis}

We first review the aggregate results across all datasets (\S\ref{sec:aggregate}) and with all metrics. We also present findings based on sentence-level analysis (\S\ref{sec:sentence}). We investigate the specific types of errors that contribute to robustness as well as those that increase undesired behavior in \S\ref{sec:errors}. Finally, in Section \S\ref{sec:divergence} we introduce the more fine-grained notion of divergence that allows us to perform interesting analysis and visualizations over the datasets.

\subsection{Aggregate Results}
\label{sec:aggregate}

Table~\ref{tab:dataset} presents the general picture of our experiments, summarizing the translation robustness across all datasets with all the metrics that we examined, and also providing basic dataset statistics. Note that the aggregate f-BLEU and f-METEOR scores in Table~\ref{tab:dataset} are calculated excluding sentences where the system exhibits robustness. 
We made this choice in order to tease apart the differences across the datasets by focusing on the problematic instances; having between 17\% and 29\% of the scores be perfect 100~f-BLEU points would obscure the analysis.
We also report average scores across all datasets (last row) as well as scores without including \lang, since the \lang dataset is significantly larger than the others.
%The robustness percentage ranges from as low as 12.4\% and up to 29\%, while f-BLEU ranges from 47 to 66 points.

\paragraph{Takeaway 1:} \textbf{Increased amounts of noise in the source degrade translation performance.} The first takeaway confirms the previous results in the literature \cite{belinkov2017synthetic,lui2019nmtnonnative}. The average number of corrections per sentence and the robustness percentage (RB) column have a Pearson's correlation coefficient $\rho=-0.82$, while both f-BLEU and f-METEOR have lower $\rho=-0.71$.

This is further outlined by the results on the \wiloc datasets. The English proficiency of the students increases from the A to B to C subsets, and the N subset is written by native English speakers. An increase in English proficiency manifests as a lower number of errors, higher robustness percentage, and larger f-BLEU scores.

\paragraph{Takeaway 2:} \textbf{The MT system generally magnifies the input noise.} This is denoted by the $\nr$ column which is larger than 1 across the board. This means that the MT system exacerbated the input noise by a factor of about $2.5$. This effect is more visible when the source noise levels are low, as in the \wiloc C and N or the \nucle datasets.

\subsection{Sentence-level Findings}
\label{sec:sentence}
\begin{figure}
    \centering
    
\pgfplotstableread[row sep=\\,col sep=&]{
nerrors & count & bleu-all & robust & bleu \\
1 & 87275 & 72.33 & 32.30 & 59.13 \\
2 & 56349 & 59.63 & 13.53 & 53.32 \\
3 & 33107 & 51.39 & 5.97 & 48.30 \\
4 & 18589 & 45.61 & 2.70 & 44.10 \\
5 & 10156 & 41.75 & 1.26 & 41.00 \\
6 & 5574 & 38.60 & 0.68 & 38.18 \\
7 & 2966 & 36.22 & 0.27 & 36.04 \\
8 & 1773 & 35.04 & 0.17 & 34.93 \\
9 & 1014 & 32.76 & 0.39 & 32.49 \\
10 & 1750 & 28.56 & 0.00 & 28.56 \\
}\posdata

\pgfplotsset{width=\textwidth/2,height=4cm,compat=1.5}
\pgfplotsset{every tick label/.append style={font=\tiny}}

\begin{tikzpicture}
    \begin{axis}[
            %axis y line*=left,
            ymajorgrids=true,
            yminorgrids=true,
            legend style={at={(0.5,1.2)},
                anchor=north,legend columns=5},
            cycle list name=black white,    
            %symbolic x coords={1,2,3,4,5,6,7,8,9,10},
            xtick={1,2,3,4,5,6,7,8,9,10},
            ytick={0,20,40,60,80},
            tick pos=left,
            %bar width=7mm, y=4mm,
            %nodes near coords,
            %nodes near coords align={vertical},
            ymin=0,ymax=60,
            xmin=0, xmax=11,
            ylabel={Percentage},
            xlabel={Number of Errors in Source},
        ]
        %\addplot+ table[x=nerrors,y=robust]{\posdata}; 
        \addplot[ybar, fill=black!50, area legend]  table[x=nerrors,y=robust,]{\posdata};
        \addplot+ table[x=nerrors,y=bleu]{\posdata};
        \legend{RB, f-BLEU}
    \end{axis}
\end{tikzpicture}

\caption{Effect of the number of errors on robustness. Robustness Percentage more than halves for each additional input sentence error, while f-BLEU on the non-robust sentences reduces linearly.}
\label{fig:number}

\end{figure}
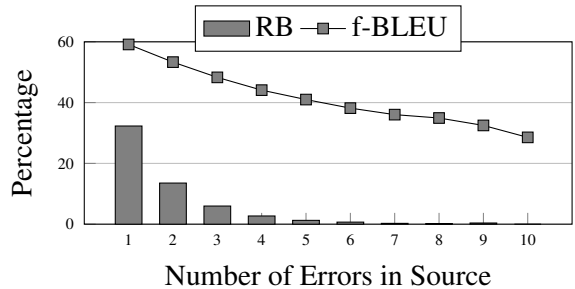
We continue our analysis focusing on instance or sentence-level factors, presenting results combining all datasets.

\paragraph{Effect of the input number of errors:} Figure~\ref{fig:number} clearly shows the compounding effect of source side errors. Each additional error reduces overall robustness by more than 50\%: from robust behavior in about 32\% of the 1-error instances, to 13\% for the 2-errors instances, to 6\% on instances with 3 errors; and so forth, to the point that the model is robust in less than 1\% of instances with more than~5 source-side errors.
The robustness drop when computed with f-BLEU is practically linear, starting from about 59 f-BLEU points when a single error is present, falling to about 28 when the source has more than 9 errors.

\iffalse % 2 Reviewers were confused and pointed out that this is irrelevant
This plot also outlines the need for multiple metrics that capture different information. For some language pairs a BLEU score of 28 would constitute a new SOTA result. In our study, though, a f-BLEU score of 28 corresponds to practically no outputs being entirely correct.\footnote{Again, BLEU and f-BLEU are not directly comparable.}
\fi

\paragraph{Effect of input length:}
\begin{figure}[t]
    \centering
    
\begin{tikzpicture}[every node/.style={inner sep=0pt}]

\node[] at (0.5,-0.2) {\small{$<$5}};
\node[] at (1.5,-0.2) {\small{$<$10}};
\node[] at (2.5,-0.2) {\small{$<$15}};
\node[] at (3.5,-0.2) {\small{$<$20}};
\node[] at (4.5,-0.2) {\small{$<$25}};
\node[] at (5.5,-0.2) {\small{$<$30}};
\node[] at (6.5,-0.2) {\small{$<$35}};

\node[] at (-0.2,0.5) {\small{1}};
\node[] at (-0.2,1.5) {\small{2}};
\node[] at (-0.2,2.5) {\small{3}};
\node[] at (-0.2,3.5) {\small{4}};
\node[] at (-0.2,4.5) {\small{5}};

\node[rotate=90] at (-0.55,2.5) {\#errors in source};
\node[rotate=0] at (3.5,-0.6) {source sentence length};
%\node (1,-1) {$<20$};

\draw[help lines] (0,0) grid (7,5);

\draw[black,fill=black,fill opacity=0.71] (0.5,0.5) circle [radius=0.16cm] ;
\draw[black,fill=black,fill opacity=0.81] (1.5,0.5) circle [radius=0.45cm] ;
\draw[black,fill=black,fill opacity=0.33] (1.5,1.5) circle [radius=0.32cm] ;
\draw[black,fill=black,fill opacity=0.14] (1.5,2.5) circle [radius=0.18cm] ;
\draw[black,fill=black,fill opacity=0.82] (2.5,0.5) circle [radius=0.46cm] ;
\draw[black,fill=black,fill opacity=0.35] (2.5,1.5) circle [radius=0.38cm] ;
\draw[black,fill=black,fill opacity=0.16] (2.5,2.5) circle [radius=0.26cm] ;
\draw[black,fill=black,fill opacity=0.07] (2.5,3.5) circle [radius=0.17cm] ;
\draw[black,fill=black,fill opacity=0.83] (3.5,0.5) circle [radius=0.35cm] ;
\draw[black,fill=black,fill opacity=0.35] (3.5,1.5) circle [radius=0.29cm] ;
\draw[black,fill=black,fill opacity=0.17] (3.5,2.5) circle [radius=0.22cm] ;
\draw[black,fill=black,fill opacity=0.08] (3.5,3.5) circle [radius=0.16cm] ;
\draw[black,fill=black,fill opacity=0.04] (3.5,4.5) circle [radius=0.12cm] ;
\draw[black,fill=black,fill opacity=0.81] (4.5,0.5) circle [radius=0.23cm] ;
\draw[black,fill=black,fill opacity=0.37] (4.5,1.5) circle [radius=0.20cm] ;
\draw[black,fill=black,fill opacity=0.16] (4.5,2.5) circle [radius=0.17cm] ;
\draw[black,fill=black,fill opacity=0.08] (4.5,3.5) circle [radius=0.14cm] ;
\draw[black,fill=black,fill opacity=0.82] (5.5,0.5) circle [radius=0.16cm] ;
\draw[black,fill=black,fill opacity=0.31] (5.5,1.5) circle [radius=0.15cm] ;
\draw[black,fill=black,fill opacity=0.14] (5.5,2.5) circle [radius=0.13cm] ;
\draw[black,fill=black,fill opacity=0.7] (6.5,0.5) circle [radius=0.10cm] ;

\end{tikzpicture}
    
    \caption{Robustness Percentage broken down by sentence length and number of source errors. The radius of each circle is proportional to the number of sentences and the opacity corresponds to RB score (darkest: RB=33\%, lowest: RB$\approx$2\%). The model is more robust with few errors regardless of sentence length.}
    \label{fig:both}
\end{figure}
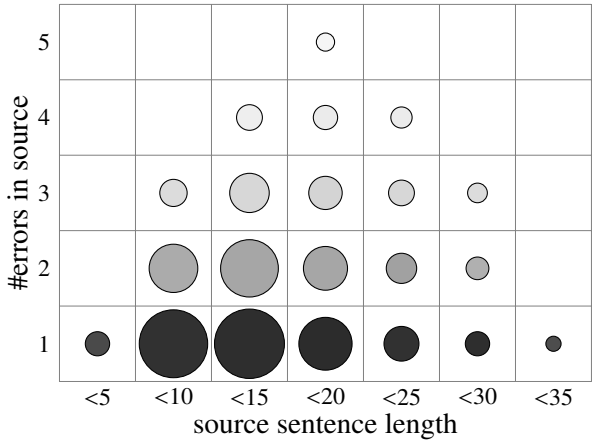 
One factor related to the number of input errors is the effect of the source sentence length. We find that there is a negative correlation between the input length and the model's robustness.
This is to be expected, as input length and the number of errors are also correlated: longer sentences are more likely to more errors, and inversely, short sentences cannot have a large number of errors.

Figure~\ref{fig:both} presents the RB score across these two factors. We bin the input sentences based on their sentence lengths and based on the number of errors in the source. We only plot bins that have a RB score of more than $1$\% (reflected in the opacity of the plot). It is clear that more errors in a source sentence lead to reduced robustness, while the sentence length is not as significant a factor.

A closer look at sentences with a single error reveals that the system is robust about~30\% of the time regardless of their length, with a slight increase in accuracy as the length increases. Longer sentences provide more context, which presumably aids in dealing with the source noise. This pattern is similar across all rows in Figure~\ref{fig:both}. 

\begin{table}[]
    \centering
    \begin{tabular}{cc|cc}
    \toprule
        
        \multicolumn{2}{c|}{Recoverable} & \multicolumn{2}{c}{Non-recoverable} \\
        Error & RB & Error & EB\\
    \midrule
        VERB-INFL & 22\% & CONJ & 3\% \\
        VERB-SVA & 22\% & OTHER & 5\% \\
        ORTH & 19\% & NOUN & 6\% \\
        VERB-FORM & 17\% & ADV & 7\% \\
        WO & 17\% & VERB & 7\% \\
    \bottomrule
    \end{tabular}
    \caption{Some errors are easier to translate correctly than others. The average error has an RB score of 11\%. We present the errors that fall out of the~$[\mu\pm2\sigma]$~range.}
    \label{tab:errors}
\end{table}
\begin{figure*}[t]
    \centering

\begin{tikzpicture}[every node/.style={inner sep=0pt}]

\node[] (c1) at (-3,-0.75) {\small{I}};
\node[right = 0.15 cm of c1] (c2) {\small{want to play with children and see their \textit{simle} all day.}};

\node[] (t10) at (-3,-1.5) {\small{\uline{Ich}}};
\node[above right = -0.23cm and 0.15 cm of t10] (t11) {\small{will}}; 
%\uline{mit} den \uline{Kindern spielen und} sie \uline{den ganzen Tag sehen.}}};

\node[below right = -0.23cm and 0.15 cm of t11] (t12) {\small{\uline{mit}}};
\node[above right = -0.23cm and 0.15 cm of t12] (t13) {\small{den}};
\node[below right = -0.23cm and 0.15 cm of t13] (t14) {\small{\uline{Kindern}}};
\node[right = 0.15 cm of t14] (t15) {\small{\uline{spielen}}};
\node[right = 0.15 cm of t15] (t16) {\small{\uline{und}}};
\node[above right = -0.23cm and 0.15 cm of t16] (t17) {\small{sie}};
\node[below right = -0.23cm and 0.15 cm of t17] (t18) {\small{\uline{den}}};
\node[below right = -0.25cm and 0.15 cm of t18] (t19) {\small{\uline{ganzen}}};
\node[below right = -0.33cm and 0.15 cm of t19] (t111) {\small{\uline{Tag}}};
\node[right = 0.15 cm of t111] (t112) {\small{\uline{sehen.}}};

\node[] (t20) at (-3,1) {\small{\uline{Ich}}};
\node[above right = -0.22cm and 0.15cm of t20] (t21) {\small{m{\"o}chte}};
\node[below right = -0.22cm and 0.15 cm of t21] (t22) {\small{\uline{mit}}};
\node[right = 0.15 cm of t22] (t23) {\small{\uline{Kindern}}};
\node[right = 0.15 cm of t23] (t24) {\small{\uline{spielen}}};
\node[right = 0.15 cm of t24] (t25) {\small{\uline{und}}};
\node[above right = -0.23cm and 0.15cm of t25] (t26) {\small{ihr}};
\node[inner sep=2pt,right = 0.15 cm of t26] (t27) {\small{L{\"a}cheln}};
\node[below right = -0.30cm and 0.15 cm of t27] (t28) {\small{\uline{den}}};
\node[below right = -0.25cm and 0.15 cm of t28] (t29) {\small{\uline{ganzen}}};
\node[below right = -0.33cm and 0.15 cm of t29] (t211) {\small{\uline{Tag}}};
\node[right = 0.15 cm of t211] (t212) {\small{\uline{sehen.}}};

\node[] (s2) at (-3,0.1) {\small{I}};
\node[right = 0.15 of s2] {\small{want to play with children and see their \textit{smiles} all day.}};

\draw[thick] (t27.south) -- (2.75,0.3);
\draw[thick] (2.75,-0.1) -- (2.75,-0.55);

\draw[] (-3.2,1.8) -- (7.5,1.8);
\node[] (cc) at (-3.8, 2) {\tiny{counts:}};
\node[below=0.3cm of cc] (pos) {\tiny{relative pos:}};
\node[] (o1) at (-3.8, -0.8) {\tiny{$x$}};
\node[] (o2) at (-3.8, -1.5) {\tiny{MT($x$)}};
\node[] (o1) at (-3.8, 0.1) {\tiny{$\tilde{x}$}};
\node[] (o2) at (-3.8, 1) {\tiny{MT($\tilde{x}$)}};

\node[above = 0.75cm of t20] (p1) {\tiny{$+0$}};
\node[above = 0.75cm of t21] (p2) {\tiny{$+1$}};
\node[above = 0.75cm of t22] (p3) {\tiny{$+0$}};
\node[above = 0.75cm of t23] (p4) {\tiny{$+0$}};
\node[above = 0.75cm of t24] (p5) {\tiny{$+0$}};
\node[above = 0.75cm of t25] (p6) {\tiny{$+0$}};
\node[above = 0.75cm of t26] (p7) {\tiny{$+1$}};
\node[above = 0.68cm of t27] (p8) {\tiny{$+1$}};
\node[above = 0.75cm of t28] (p9) {\tiny{$+0$}};
\node[above = 0.82cm of t29] (p11) {\tiny{$+0$}};
\node[above = 0.75cm of t211] (p12) {\tiny{$+0$}};
\node[above = 0.75cm of t212] (p13) {\tiny{$+0$}};

\node[below=0.3cm of p1] (pp1) {\tiny{$-7$}};
\node[below=0.3cm of p2] (pp2) {\tiny{$-6$}};
\node[below=0.3cm of p3] (pp3) {\tiny{$-5$}};
\node[below=0.3cm of p4] (pp4) {\tiny{$-4$}};
\node[below=0.3cm of p5] (pp5) {\tiny{$-3$}};
\node[below=0.3cm of p6] (pp6) {\tiny{$-2$}};
\node[below=0.3cm of p7] (pp7) {\tiny{$-1$}};
\node[below=0.3cm of p8] (pp8) {\textcolor{red}{\tiny{$0$}}};
\node[below=0.3cm of p9] (pp9) {\tiny{$1$}};
\node[below=0.3cm of p11] (pp11) {\tiny{$2$}};
\node[below=0.3cm of p12] (pp12) {\tiny{$3$}};
\node[below=0.3cm of p13] (pp13) {\tiny{$4$}};

\end{tikzpicture}

    \caption{The procedure of computing \textit{divergence} over a quadruple $(\xvec,\yvec,\xtvec,\ytvec)$. Each token in output $y$ not in the desired output $\tilde{y}$ is considered a divergent token (\uline{underlined}=matching). The x-axis is centered around the token $\tilde{y}_k$ that aligns to the edit $x_i$*$\rightarrow\tilde{x}_j$. The counts describe the caused divergence relative to the expected error's position.}
    \label{fig:dive}
\end{figure*}

\subsection{Error-level Analysis}
\label{sec:errors}

In this section we aim to study and identify the error types from which the NMT system is able to recover, or not.
To avoid the compounding effects of multiple source-side errors, we restrict this analysis to sentences that have a single error.

We have already discussed in Section~\ref{sec:aggregate} how the NMT system is robust on about 20\% of the instances across all corpora.
By selecting those instances and computing basic error statistics on them, we find that the average error is recoverable about 11\% of the time ($\mu=0.11$). 
Table~\ref{tab:errors} presents the errors that are harder or easier to translate correctly. We choose to present the errors that are at the bottom and top, respectively, of the ranking of the errors, based on the average RB score that their corresponding test instances receive.

The non-recoverable errors on the right side of Table~\ref{tab:errors} are mostly semantic in nature: all five of them correspond to instances where a semantically wrong \textit{word} was used.\footnote{We refer the reader to \citet{bryant-felice-briscoe:2017:Long} for a complete list of the error type abbreviations.}
Correcting and even identifying these types of errors is difficult even in a monolingual setting as world knowledge and/or larger (document/discourse) context is needed. One could argue, in fact, that such errors are not \textit{grammatical}, i.e. the source sentence is fluent. Furthermore, one could form a solid argument for not wanting/expecting an \textit{MT} system to alter the semantics of the source. The MT system's job is exactly to accurately convey the semantics of the source sentence in the target language.

However, there are errors where the intended meaning is clear but ungrammatically executed, as in Table~\ref{tab:errors}'s left-side errors. There are three plausible (likely orthogonal, but we leave such analysis for future work) reasons why these errors are easier than average to correctly translate:

\paragraph{1. Self-attention.} The encoder's final representations are computed through multiple self-attention layers, resulting in a representation heavily informed by the whole source context.
The VERB-INFL, VERB-SVA, and VERB-FORM error categories (all related to morphology and syntactic constraints) apply to edits that subword modeling combined with self-attention would alleviate. Consider the example of the verb inflection (VERB-INFL) error \textit{danceing*/dancing}. The segmentation in the erroneous and the corrected version is \textit{dance$\vert$ing} and \textit{danc$\vert$ing} respectively. In both cases, the morpheme that denotes the inflection is the same. 
Verb form (VERB-FORM) errors, on the other hand, typically involve infinitive, to-infinitive, gerund, or participle forms.
It seems that in those cases the self-attention component is able to use the context to recover, especially because, as in the VERB-INFL example, the stem of the verb will most likely be the same.

Also, apart from the positional embeddings, no other explicit word order information is encoded by the architecture (unlike recurrent architectures focused on by all previous work, which by construction keep track of word order). We suggest that the self-attention architecture makes word order errors (WO errors are strictly defined as exact match tokens wrongly ordered, e.g. \textit{know already*/already know}) easier to recover from.

\paragraph{2. The extensive use of back-translation.} The SOTA model that we use has been trained on massive amounts of back-translated data, where German monolingual data have been translated into English. The integral part is that English sources were \textit{sampled} from the De-En model, instead of using beam-search to generate the most likely output. This means that the model was already trained on a fair amount of source-side noise, which would make it more robust to such perturbations \cite{belinkov2017synthetic,lui2019nmtnonnative,singh2019improving}.

Although we do not have access to the back-translated parallel data that \citet{edunov2018} used, we suspect that translation errors are fairly common and therefore more prevalent in the final training bitext, making the model more robust to such noise.
Current English-to-German SOTA systems
%\footnote{like the one used for back-translation}
might not have issues with translating noun phrases, coordinated verbs, or pronoun number, but they still struggle with compound generation, coreference, and verb-future generation \cite{bojar2018proceedings}.

\paragraph{3. Data preprocessing and subword-level modeling.}
It is worth noting that ERRANT limits the orthography (ORTH) error category to refer to edits involving casing (lower$\leftrightarrow$upper) and whitespace changes.
Our model, as most of the SOTA NMT models, is trained and operates at the subword level, using heuristic segmentation algorithms like Byte Pair Encoding (BPE) \cite{sennrich2016}, that are learned on clean truecased data. Truecasing is also a standard preprocessing step at inference time, hence dealing with casing errors.
The BPE segmentation also has the capacity to deal with whitespace errors. For example, the incorrect token ``weatherrelated" gets segmented to \textit{we$\vert$a$\vert$ther$\vert$related}. Although imperfect (the segment's segmentation with proper whitespacing is \textit{we$\vert$a$\vert$ther related}), the two segmentations agree for~3/4 tokens. 
%while we expect the representations of {\small\textit{ther}} and {\small\textit{ther$\vert$}} to be quite similar either way.
Most previous work
e.g. \cite{belinkov2017synthetic}
has focused on character-level modeling using compositional functions across characters to represent tokens, which are by construction more vulnerable to such errors. \normalfont

\begin{figure*}[t!]
    \centering
    \small\addtolength{\tabcolsep}{-5pt}
    \begin{tabular}{@{}C{.32\textwidth}@{}|C{.32\textwidth}|@{}C{.32\textwidth}@{}}
    \toprule
        %\multicolumn{3}{c}{Errors with the smallest mean also have the highest variance}  \\
        R:ORTH & M:VERB:FORM & U:ADJ\\
        \tiny $\mu$: -1.9 \ $\sigma$: 5.9 \ $\gamma_1$: -2.1&
        \tiny $\mu$: -1.4 \ $\sigma$: 7.6 \ $\gamma_1$: -1.2 &
        \tiny $\mu$: -1.1 \ $\sigma$: 6.1 \ $\gamma_1$: -1.1\\
        \includegraphics[width=0.31\textwidth]{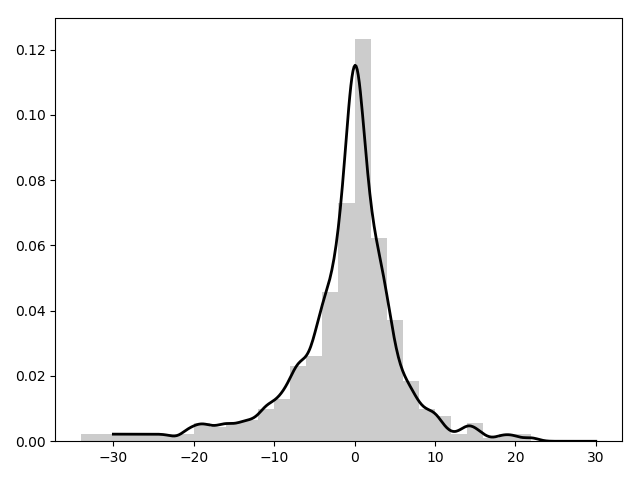} &
        \includegraphics[width=0.31\textwidth]{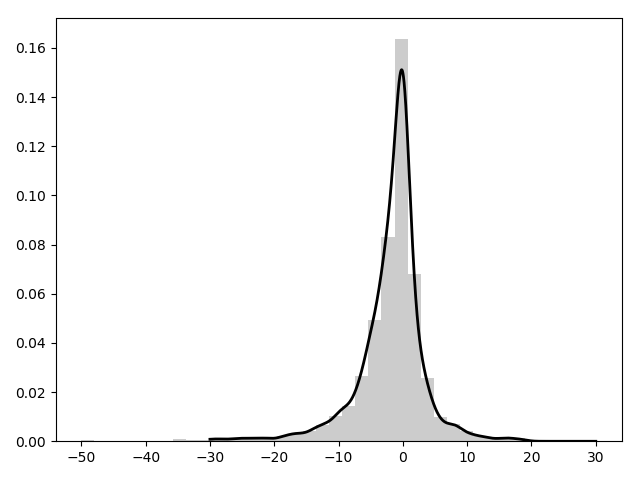} &
        \includegraphics[width=0.31\textwidth]{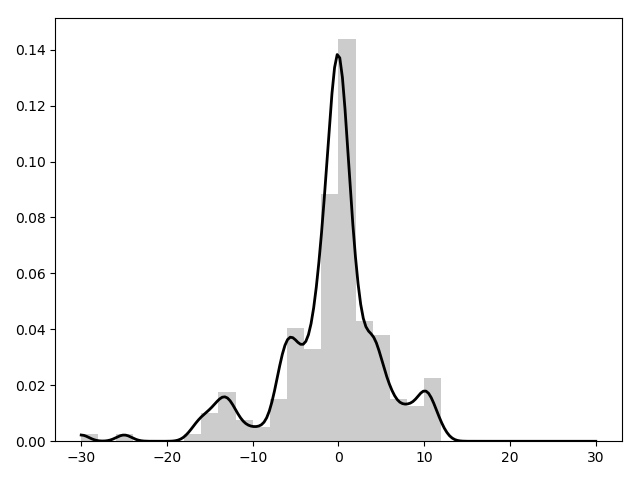} \\
        \textbf{lowest mean}, large variance, & 
        low mean, \textbf{largest variance},&
        low mean, large variance,\\
         \textbf{most negative skewness} &  negative skewness & negative skewness  \\
        \midrule
        
        %\multicolumn{3}{c}{Errors with the largest mean}\\
        U:CONTR & R:WO & U:CONJ \\
        \tiny $\mu$: 1.6 \ $\sigma$: 3.8 \ $\gamma_1$: 0.1 &
        \tiny $\mu$: 1.8 \ $\sigma$: 4.9 \ $\gamma_1$: 1.0 &
        \tiny $\mu$: 2.3 \ $\sigma$: 5.6 \ $\gamma_1$: -0.2 \\
        \includegraphics[width=0.31\textwidth]{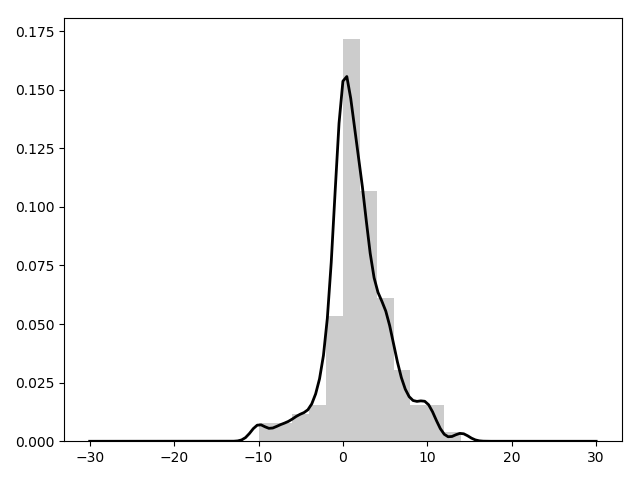} &
        \includegraphics[width=0.31\textwidth]{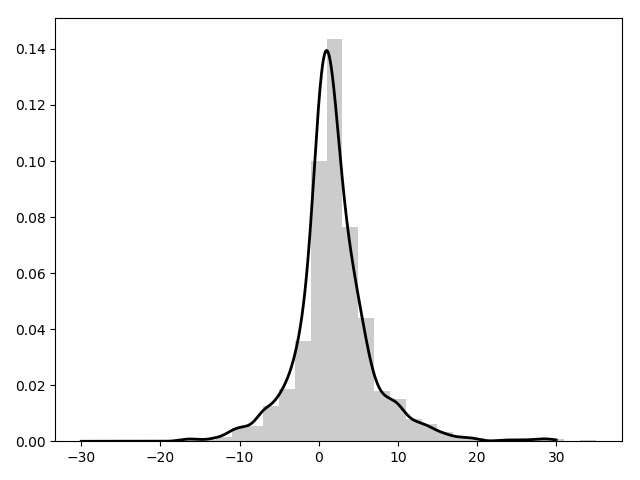} &
        \includegraphics[width=0.31\textwidth]{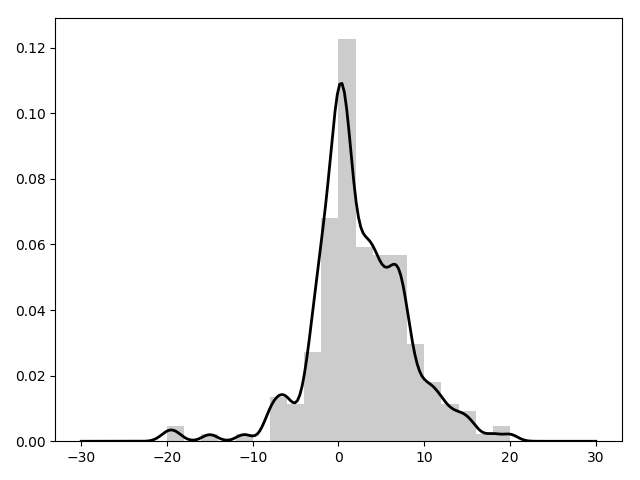} \\
        large mean, \textbf{smallest variance},&
        large mean, small variance, &
        \textbf{largest mean}, large variance,\\
         no skewness  & positive skewness &  no~skewness \\
        \midrule

        %\multicolumn{3}{c}{Errors with statistically significant skewness}\\
        U:PART & R:SPELL & R:CONTR\\
        \tiny $\mu$: 0.6 \ $\sigma$: 5.7 \ $\gamma_1$: -1.9 &
        \tiny $\mu$: 0.5 \ $\sigma$: 5.6 \ $\gamma_1$: 5.9 &
        \tiny $\mu$: 1.1 \ $\sigma$: 5.6 \ $\gamma_1$: 2.1 \\
        \includegraphics[scale=0.3]{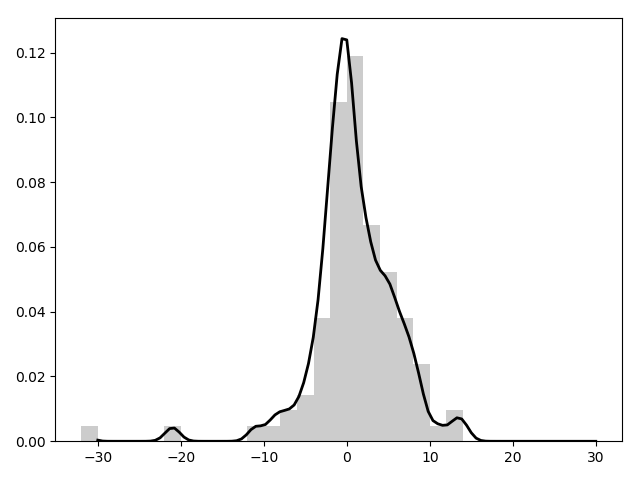} &
        \includegraphics[scale=0.3]{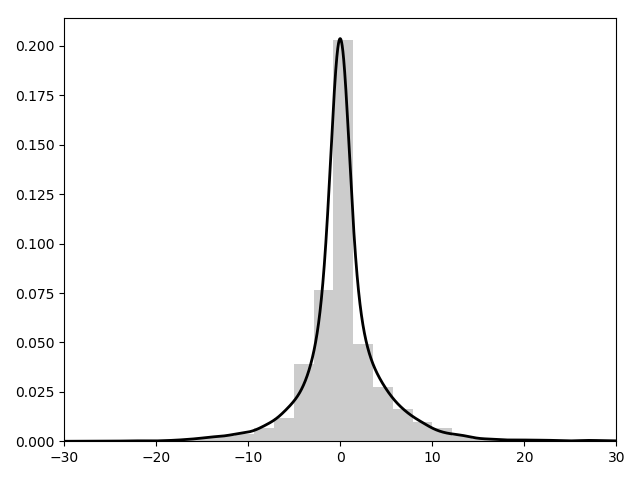} &
        \includegraphics[scale=0.3]{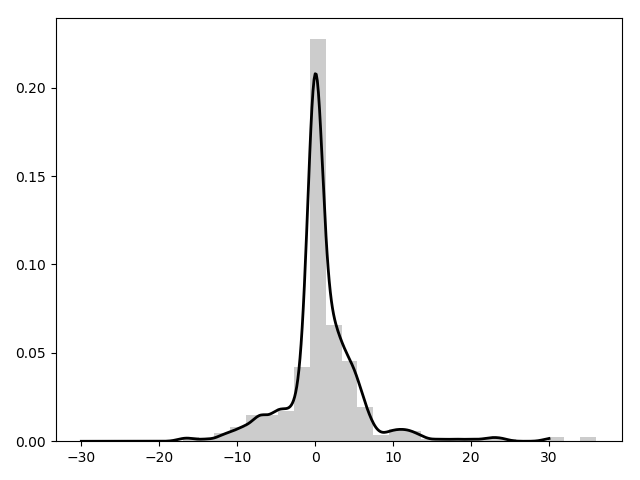} \\
        close-to-zero mean, large variance, &
        almost-zero mean, large variance, & 
        large variance, \\
        large negative skewness & \textbf{largest positive skewness}  & large positive skewness\\
        \bottomrule
    \end{tabular}
    \normalfont
    \caption{Some interesting errors with statistics on their \textit{divergence distribution}. Some errors (negative mean and skewness: \small{R:ORTH, M:VERB:FORM, U:ADJ}) affect the left context of their translation more, while others affect their right translation context (positive mean and skewness \small{R:WO, U:CONJ}). Errors might affect a small neighborhood (low variance: \small{U:CONTR, R:WO}) or a larger part of the translation (high variance: \small{M:VERB:FORM, U:ADJ, R:CONTR}).}
    \label{tab:examples}
\end{figure*}
\begin{figure*}[t!]
    \centering
    \small\addtolength{\tabcolsep}{-5pt}
    \begin{tabular}{@{}C{.245\textwidth}@{}|@{}C{.245\textwidth}@{}|@{}C{.245\textwidth}@{}|@{}C{.245\textwidth}@{}}
    \toprule
        %\multicolumn{3}{c}{Errors with the smallest mean also have the highest variance}  \\
        first quartile \ \ \ \ \ \ \ \ \ \ \ \ \ ($i^* < 0.25\vert x \vert$) & second quartile ($0.25\vert x \vert \leq  i^* < 0.50\vert x \vert$) & third quartile ($0.50 \vert x \vert \leq i^* < 0.70\vert x \vert$)& fourth quartile ($0.75 \vert x \vert \leq i^* <\vert x \vert$)\\
        $\mu$: 3.8 \ $\sigma$: 5.6 \ $\gamma_1$: 3.3 &
        $\mu$: 1.6 \ $\sigma$: 4.1 \ $\gamma_1$: 1.2 &
        $\mu$: -0.2 \ $\sigma$: 3.6 \ $\gamma_1$: -1.1&
        $\mu$: -2.9 \ $\sigma$: 4.8 \ $\gamma_1$: -2.5\\
        \includegraphics[width=0.24\textwidth]{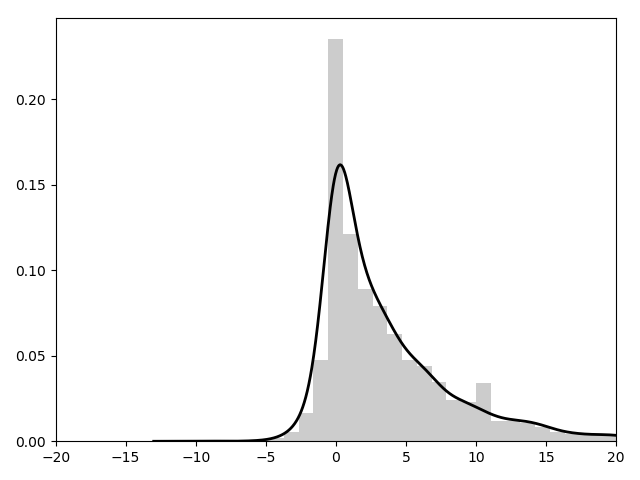} &
        \includegraphics[width=0.24\textwidth]{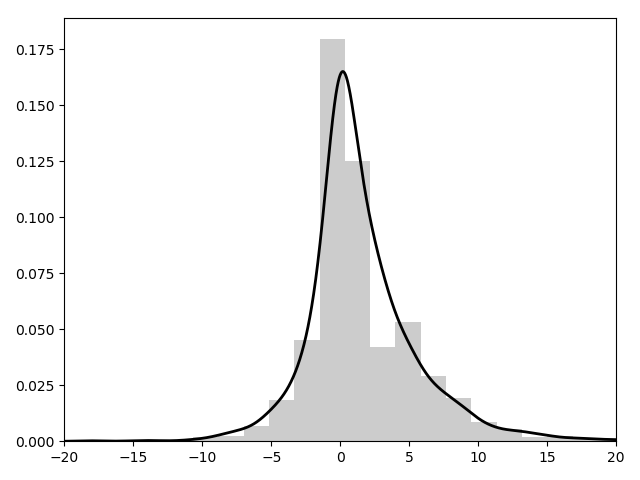} &
        \includegraphics[width=0.24\textwidth]{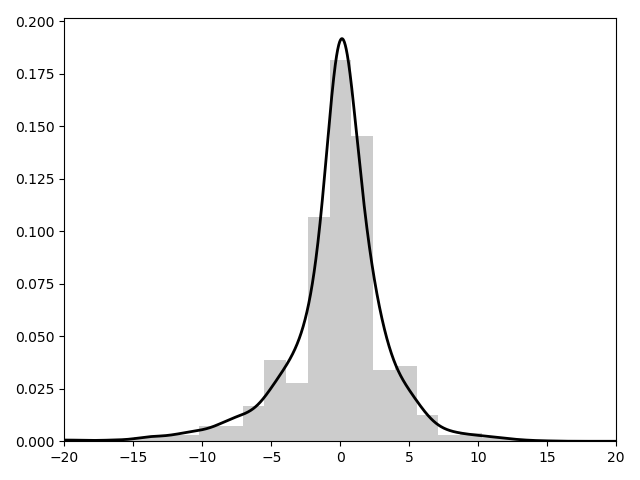}
        &
        \includegraphics[width=0.24\textwidth]{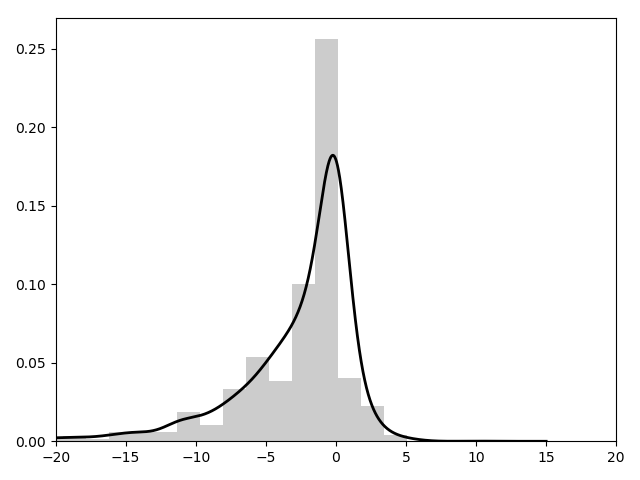}\\
        \bottomrule
    \end{tabular}
    \normalfont
    \caption{Divergence Distributions for single source error instances per the error's location quartile.}
    \label{tab:len}
\end{figure*}

\subsection{Divergence}
\label{sec:divergence}

We introduce a method for computing a \textit{divergence distribution}. 
%This method will allow us to visualize the effect of source side errors on output sentences.
Computing \textit{divergence} requires a quadruple of $(\mathbf{x},\tilde{\mathbf{x}},\mathbf{y},\tilde{\mathbf{y}})$. %Even though such data are hard to come by outside our scenario, as they require source side correction annotations, we find that they allow for interesting visualizations.
We will focus on instances where $\xvec$ and $\xtvec$ differ only with a single edit, as a simple working example.

\paragraph{Process:}
Given a source side sentence pair $\mathbf{x}$ and $\mathbf{\tilde{x}}$ with a single grammatical error, it is trivial to identify the position $i^*$ of the correction in $\tilde{\mathbf{x}}$, since we work on corpora pre-annotated with grammatical edits at the token level.
Also, using traditional methods like the IBM Models \cite{brown1993mathematics} and the GIZA++ tool, makes it easy to obtain an alignment between $\mathbf{x}$ and $\mathbf{y}$, as well as between $\tilde{\mathbf{x}}$ and $\tilde{\mathbf{y}}$. We use the alignment variable $\alpha_j = i$ to denote that the target word $y_j$ is aligned to source position $i$, and equivalently the variables $\tilde{\alpha}$ for the corrected source pair.
We denote as $k^*$ the target position that aligns to the source-side correction, such that $\tilde{\alpha}_{k^*} = i^*$.

We define the set of divergent tokens $\mathcal{Y*}$ as the set of tokens of $\mathbf{y}$ that do not appear in $\tilde{\mathbf{y}}$:
%{
%\setlength{\abovedisplayskip}{5pt}
%\setlength{\belowdisplayskip}{5pt}
\[
\mathcal{Y^*} = \{y_j \mid y_j \not\in \tilde{\mathbf{y}}\}.
\]
%}
Now, we use all the previous definitions to define the set $\mathcal{P}$ of target divergent positions for a quadruple $(\mathbf{x},\tilde{\mathbf{x}},\mathbf{y},\tilde{\mathbf{y}})$ as the set of target-side positions of the tokens that are different between $\mathbf{y}$ and $\tilde{\mathbf{y}}$, but \textit{relative} to the position of the target-side token that aligns to the source-side correction:

\[
\mathcal{P}(\mathbf{x},\tilde{\mathbf{x}},\mathbf{y},\tilde{\mathbf{y}}) = \{ j-k*, \forall y_j \in \mathcal{Y^*} \}.
\]

We provide an illustration of this process for a single-error example in Figure~\ref{fig:dive}. The correction \textit{simle*/smiles} is aligned to the word $y_7$ (L{\"a}cheln) in the reference target, so the center of the distribution is moved to $k^* = 7$. 
For the rest of the positions in the target reference $\mathbf{\tilde{y}}$, we simply update the counts based on whether the word $\tilde{y}_j$ is present in $\mathbf{y}$. 
The final step is collecting counts across all instances for all the relative divergent positions and analyzing the effect of a source-side error on the target sentence.

Essentially, we expect some source-side errors to have a very local effect on the translation output, which would translate in divergence distributions with low variance (since we center the distribution around $k^*$). Other source-side errors might cause larger divergence as they might affect the whole structure of the target sentence.

In the Figure~\ref{fig:dive} example, the only difference between $\xvec$ and $\xtvec$ is a single word towards the end of the sentence, but the outputs $\yvec$ and $\ytvec$ diverge on three words.
One of them is 6 words away (before) from where we would have expected the divergence to happen (in relative position~0).

After collecting divergence counts for each instance, we can visualize their distribution and compute their descriptive statistics. We focus on the mean $\mu$, standard deviation $\sigma$, and the skewness of the distribution as measured by Pearson's definition, using the third standardised moment, defined as: 
\[
\gamma_1 = \mathbb{E}\left[\left( \frac{X-\mu}{\sigma}\right)^3\right].
\]

Across all datasets and errors, the distribution of the divergence caused by single errors in the source has a mean $\mu_{all}$=$0.7$, standard deviation $\sigma_{all}$=$5.1$, and a slight positive skewness with $\gamma_{1_{all}}$=$0.8$. This means that the average error affects its general right context, in a $\pm5$ word neighborhood. 

In Figure~\ref{tab:examples} we present several of the errors with the most interesting divergence statistics. Some errors heavily affect their left (e.g. R:ORTH) or right context (U:CONJ). Also, some errors affect a small translation neighborhood as denoted by the low variance of their divergence distribution (e.g. U:CONTR). On the other hand, verb form errors (M:VERB:FORM) have the potential to affect a larger neighborhood: this is expected because English auxiliary verb constructions (e.g. "have eaten X") often get translated to German V2 constructions with an auxiliary verb separated from a final, non-finite main verb (e.g. ``habe X gegessen").

In Figure~\ref{tab:len} we present the divergence distributions across the sentence quartiles where the error appears. We find that errors in the sentence beginning (1\textsuperscript{st} quartile) severely affect their right context. Errors towards the end of the sentence (4\textsuperscript{th} quartile) affect their left context. Interestingly, we observe that mid-sentence errors (2\textsuperscript{nd}, 3\textsuperscript{rd} quartiles) exhibit much lower divergence variance than errors towards the sentence's edges.

\section{Limitations and Extensions}

A major limitation of our analysis is the narrow scope of our experiments. We solely focused on a single language pair using a single MT system. Whether different neural architectures over other languages would lead to different conclusions remains an open question. The necessary resources for answering these questions at scale, however, are not yet available. We were limited to English as our source side language, as the majority of the datasets and research works in GEC are entirely English-centric. Perhaps small-scale GEC datasets in Estonian~\cite{rummo-praakli-2017} and Latvian~\cite{deksne2014error} could provide a non-English testbed. One would then need error labels for the grammatical edits, so if such annotations are not available, an extension of a tool like ERRANT to these languages would also be required. One should also be careful in the decision of what (N)MT system to test, as using a low-quality translation system would not produce meaningful insights.

Another limitation is that our metrics do not capture whether the changes in the output are actually grammatical errors or not. In the example in Figure~\ref{fig:dive}: the German words ``m{\"o}chte" and ``will" that we identified as divergent are practically interchangeable. Therefore, the NMT model is technically not wrong outputting either of them and it is indeed generally possible that differences between~$y$ and~$\tilde{y}$ are just surface-level ones. The inclusion of f-METEOR as a robustness metric could partially deal with this issue, as it would not penalize such differences.
We do believe it is still interesting, though, that a single source error can cause large perturbations in the output, as in the case of errors with large variance in their divergence distribution.
Nevertheless, an extension of our study focusing on the grammatical qualities of the MT output would be exciting and automated tools for such analysis would be invaluable (i.e. MT error labeling and analysis tools extending the works of \citet{zeman2011addicter}, \citet{logacheva2016marmot}, \citet{popovic2018error}, or \citet{neubig19naacl}).

A natural next research direction is investigating how to use our reference-less evaluation metrics in order to create a more robust MT system. For instance, one could optimize for f-BLEU or any of the other reference-less measures that we proposed, in the same way that an MT system is optimized for BLEU (either by explicitly using their scores through reinforcement learning or by simply using the metric as an early stopping criterion over a development set). \citet{cheng-EtAl:2018:Long} recently proposed an approach for training more robust MT systems, albeit in a supervised setting where noise is injected on parallel data, and the proposed solutions of \citet{belinkov2017synthetic} and \citet{lui2019nmtnonnative} fall within the same category. However, no approach has, to our knowledge, used GEC corpora for training MT systems robust to grammatical errors. In any case, special care should be taken so that any improvements on translating ungrammatical data do not worsen performance on clean ones.

\section{Conclusion}

In this work, we studied the effects of grammatical errors in NMT.
We expanded on findings from previous work, performing analysis on real data with grammatical errors using a SOTA system. 
With our analysis we were able to identify classes of grammatical errors that are recoverable or irrecoverable.
Additionally, we presented ways to evaluate a MT system's robustness to noise without access to gold references, as well as a method for visualizing the effect of source-side errors to the output translation.
Finally, we discussed the limitations of our study and outlined avenues for further investigations towards building more robust NMT systems.

\section*{Acknowledgements}
The author is grateful to the anonymous reviewers, Kenton Murray, and Graham Neubig for their constructive and insightful comments, as well as to Gabriela Weigel for her invaluable help with editing and proofreading the final version of this paper. This material is based upon work generously supported by the National Science Foundation under grant~1761548.

\bibliographystyle{acl_natbib}
\bibliography{References}

\end{document}